\documentclass[conference]{IEEEtran}
\IEEEoverridecommandlockouts
\usepackage[utf8]{inputenc}
\usepackage{csquotes}
\usepackage{listings}
\usepackage{cite}
\usepackage{xcolor}
\definecolor{light-gray}{gray}{0.95}

\usepackage{makecell}
\usepackage{multirow}     
\usepackage{mdframed}
\usepackage[breakable, many]{tcolorbox} %
\usepackage{balance}
\newtcolorbox{answerbox}{
    breakable,
    boxrule=0pt,
    leftrule=2pt,
    fontupper=\linespread{0.9}\small,
    top=1pt, 
    bottom=1pt, 
    left=2pt,  
    right=2pt,
}
\usepackage{url}
\usepackage{graphicx}
\lstdefinestyle{customlisting}{
    basicstyle=\linespread{0.9}\ttfamily\scriptsize,
    backgroundcolor=\color{white},
    frame=none,
    breaklines=true,
    breakindent=0pt,
    showstringspaces=false,
    aboveskip=0pt,      
    belowskip=0pt,    
    abovecaptionskip=2pt,
    belowcaptionskip=2pt,
}
\tcbset{
    customstyle/.style={
        colframe=black,       
        colback=white,  
        boxrule=0.2mm,        
        breakable,            
        left=8pt, right=8pt,  
        top=2pt, bottom=2pt,  
        boxsep=0pt            
    }
}
\usepackage{tabularx}
\usepackage{booktabs}
\usepackage{makecell} 
\usepackage{adjustbox}
\usepackage{enumitem} 
\usepackage{tcolorbox}
\usepackage{verbatim} 
\usepackage{color}
\usepackage{colortbl}

\usepackage{amsmath,amssymb,amsfonts}
\usepackage{algorithmic}
\usepackage{textcomp}
\usepackage{threeparttable}
\usepackage{fancyhdr}
\usepackage{dblfloatfix}
 \usepackage{subcaption}
\usepackage{wrapfig}
\usepackage{scrhack}
\usepackage{lstautogobble}
\usepackage{tikz}
\usepackage{pgfplots}
\usepackage{pgfplotstable}
\usepackage{caption}
\usepackage{lscape}
\usepackage{afterpage}
\usepackage{float}

\def\BibTeX{{\rm B\kern-.05em{\sc i\kern-.025em b}\kern-.08em
    T\kern-.1667em\lower.7ex\hbox{E}\kern-.125emX}}
  
\newcommand{\mytilde}{\raise.17ex\hbox{$\scriptstyle\mathtt{\sim}$}}

\definecolor{cgray}{RGB}{225, 225, 225} 

\definecolor{bluebars}{HTML}{638EC6}
\definecolor{pinkbars}{HTML}{F26E8D}

\newcommand{\mybar}[2][bluebars]{%
  {\color{#1}\rule{#2pt}{8pt}}\,\textbf{#2\%}%
}

\newcommand\blfootnote[1]{%
  \begingroup
  \renewcommand\thefootnote{}\footnote{#1}%
  \addtocounter{footnote}{-1}%
  \endgroup
}

\def\BibTeX{{\rm B\kern-.05em{\sc i\kern-.025em b}\kern-.08em
    T\kern-.1667em\lower.7ex\hbox{E}\kern-.125emX}}
\begin{document}

\title{How Well Can AI Generate Backlogs from App Mockups?}

\author{\IEEEauthorblockN{Andrea Lezcano Airaldi}
\IEEEauthorblockA{\textit{Department of Informatics} \\
\textit{National University of the Northeast}\\
Corrientes, Argentina \\
alezcano@exa.unne.edu.ar}
\and
\IEEEauthorblockN{Lourdes Romera}
\IEEEauthorblockA{\textit{Department of Informatics} \\
\textit{National University of the Northeast}\\
Corrientes, Argentina \\
louromeera@gmail.com}
\and
\IEEEauthorblockN{Walid Maalej}
\IEEEauthorblockA{\textit{Hasso Plattner Institute} \\
\textit{University of Potsdam}\\
Potsdam, Germany \\
walid.maalej@hpi.de}
}

\maketitle

\blfootnote{\footnotesize © 2026 IEEE. Personal use of this material is
permitted. Permission from IEEE must be obtained for all other uses, in any
current or future media, including reprinting/republishing this material
for advertising or promotional purposes, creating new collective works,
for resale or redistribution to servers or lists, or reuse of any
copyrighted component of this work in other works.}

\begin{abstract}
Creating sprint backlogs requires considerable effort, as items such as epics, user stories, and tasks can be missed or inconsistently specified.
We propose a multimodal approach to support backlog generation from visual app mockups, an artifact available at early project stages.
We evaluate three prompting strategies on GPT-4o: a zero-shot baseline, Compositional Chain-of-Thought (CCoT) for vision-language reasoning, and a persona-driven prompt.
We study seven app development projects across two countries and interview developers about the results. 
Overall, we observed that the baseline prompt favours recall over precision, whereas CCoT is more balanced, achieving average F1 scores of 52–66\% for epics and user stories.
Tasks were more challenging to generate accurately. 
Precision gains were most consistent when adding architectural context, particularly for backend tasks (precision gains up to 35\%).
Interviews with developers revealed that up to 26\% of false positives were still considered useful, reflecting the creative and open-ended nature of backlog creation.
To capture this, we propose a new measure called Revised Recall, which complements ground-truth evaluation with developer assessments.
Our findings suggest that hybrid prompting with architectural context can assist backlog generation from early mockups, though results vary by item type and developer oversight remains necessary.
\end{abstract}


\section{Introduction}
\label{sec:introduction}

A precise and complete backlog is key to software project success. 
Yet backlog items including epics, user stories, and tasks~\cite{montgomery_alternative_2022} can be missed~\cite{fernandez_naming_2017} 
and are prone to ambiguity and inconsistency~\cite{VanCan2025}. 
Empirical studies show that backlog items are often described inconsistently, merge multiple requirements, or omit rationale and technical details~\cite{VanCan2025}. This can hinder planning and implementation. 
Requirements specification via epics and user stories is among the most challenging software engineering activities~\cite{fernandez_naming_2017}. 
And, refining user stories into detailed tasks is time-consuming and may introduce variations that affect sprint velocity and software quality~\cite{Müter2019}.

\begin{figure}[b]
    \centering
    \includegraphics[width=\columnwidth]{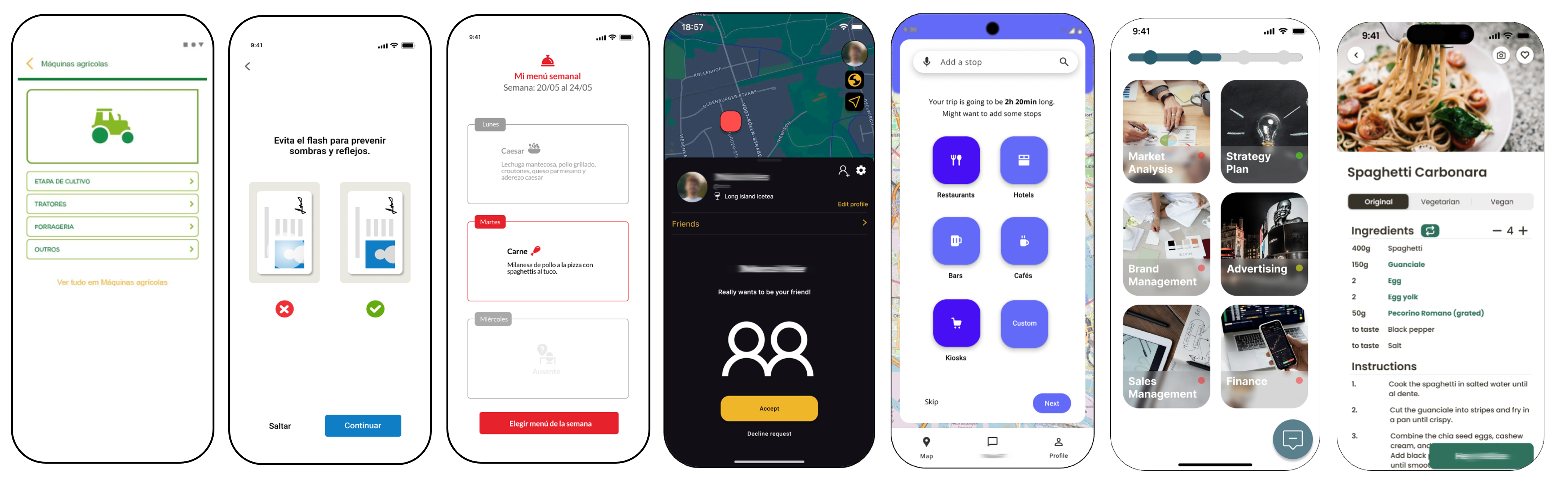}
    \caption{Examples of app mockups used as input for our study.}
    \label{fig:mockups}
\end{figure}

Recent advances in generative artificial intelligence (GenAI) have demonstrated great potential for generating code \cite{Jiang:TOSEM:2025} and other project artefacts including textual requirements \cite{Rahman2024} and user interfaces  \cite{Wei2024b}. 
Yet most existing approaches focus on textual inputs, and  to our knowledge, none investigate entire backlogs.  
Because many agile teams (particularly in app development) rely on easy-to-create visual mockups~\cite{rivero2014} as the example on Fig.~\ref{fig:mockups} shows, automatically  translating these visuals into backlog items could reduce manual effort and support more systematic backlog creation.

In this paper, we study the use of Multimodal Foundation Models, specifically GPT-4o, to support backlog item generation from visual mockups. 
We focus on three types of items: epics, user stories/scenarios, and tasks. 
\textbf{Epics} are broad narratives that group related user stories and are sometimes called requirements or features. 
\textbf{User stories} are short, structured descriptions of user needs, that capture system interactions from an end-user perspective. 
Similarly, \textbf{scenarios} narrate how users interact with a system to achieve specific goals. 
Finally, \textbf{Tasks} are actionable items derived from user stories and scenarios~\cite{Müter2019}, are often assigned to team members, and tied to system architecture or implementation details.

We evaluate three prompting strategies: a simple baseline, a compositional chain-of-thought (CCoT) optimized for vision-language reasoning~\cite{Mitra2023}, and a persona-driven prompt incorporating user perspectives~\cite{Khanh2017}. 
We also examine whether adding architectural context affects task generation, since tasks often depend on implementation details not visible in mockups. 
Finally, we assess perceived usefulness of generated backlogs through qualitative feedback from development teams. Accordingly, the research questions are:
\begin{itemize}
\item \textbf{RQ1}: How do different \underline{prompting strategies} perform in generating backlog items?
\item \textbf{RQ2}: How does \underline{architectural context} impact backlog generation?
\item \textbf{RQ3}: How do development teams perceive the \underline{usefulness} of backlog generation and the generated backlog items?
\end{itemize}

Our results indicate that the baseline prompt achieves a relatively high recall for epics and user stories, whereas task generation remains limited.
The CCoT prompt provides a more balanced trade-off between recall and precision, particularly in non-hierarchical backlogs. 
The persona prompt captures user-specific aspects but typically lowers recall. 
Including architectural context often increases task generation performance, narrowing the gap with epics and user stories. 
However, basic architectural information can introduce misaligned details, especially for higher-level tasks.
Across teams, up to 26\% of items initially labelled as \textit{false positives} were considered useful. 
This highlights the limitation of ground-truth-based evaluations that might lack items developers initially missed, and underscores the potential of GenAI to support developers in creative, ``open-ended'' tasks as backlog creation through iterative conversational interactions \cite{Wei:Software:2025}.

We present the research methodology, including the data collection and analysis in Section~\ref{sec:methodology}. Section~\ref{sec:results} reports the results along the RQs while Section~\ref{sec:discussion} discusses the findings, implications, and limitations. Section~\ref{sec:relwork} reviews related work, and Section~\ref{sec:conclusion} concludes the paper.

\section{Research Methodology}
\label{sec:methodology}
\begin{figure}[]
    \centering
    \includegraphics[width=\columnwidth]{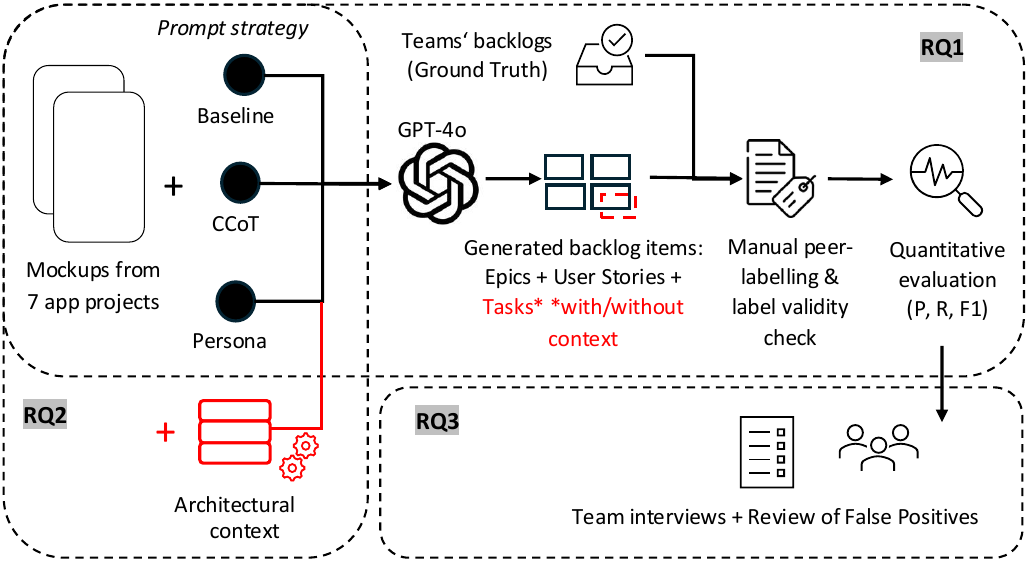}
    \caption{Overview of the research methodology.}
    \label{fig:method}
\end{figure}

We studied mockups and backlogs from seven app development projects and interviewed their developers.
A mockup is an individual UI screen (Fig.~\ref{fig:mockups}) representing a distinct interface state or interaction.
Using OpenAI's GPT-4o~\cite{OpenAI}, we generated backlog items from mockups and compared them to the original backlogs via manual assessment.
For RQ1, we evaluated prompts quantitatively using precision, recall, and F1 against the ground truth.
For RQ2, we extended this analysis by adding architectural context for task generation.
For RQ3, we conducted team interviews and reviewed false positives to identify relevant but undocumented items in the ground-truth.
Fig.~\ref{fig:method} summarizes the methodology  detailed in the following.

\subsection{Prompt Design}
\label{subsec:prompt_selection}
We implemented the three prompting strategies introduced in Section \ref{sec:introduction}, following prompt engineering best practices \cite{White2023}, providing an input, explicit task instructions, and a structured output format. 
The CCoT and Persona prompts extend the baseline by incorporating additional model reasoning steps. 
Below we describe each prompting strategy. All corresponding templates are available in our replication package\footnote{\url{https://tinyurl.com/rep-package}}.

\noindent\textbf{Baseline Prompt.} 
A zero-shot reference prompt without additional guidance.
It consists of an input of app mockups without textual descriptions, the target backlog item type, and the output format (e.g., \texttt{As \{user\}, I want \{feature\} so that \{benefit\}} for user stories). Full templates are in the replication package.
    
\noindent\textbf{Compositional Chain-of-Thought.}
Chain-of-Thought (CoT) prompting guides LLMs to produce intermediate reasoning steps before the final answer. 
By breaking complex questions into smaller steps, CoT has been shown to improve output quality in domains such as mathematics, programming, and general problem solving~\cite{Wang2022, Li:TOSEM:2025}. 
Extensions include self-consistency~\cite{Wang2022}, which samples multiple reasoning chains and selects the most coherent, and Tree-of-Thought~\cite{Yao} and Graph-of-Thought, which explore non-linear paths and revisit prior steps.
Compositional Chain-of-Thought (CCoT) introduces a structured intermediate representation (a scene graph) to guide reasoning over visual artefacts~\cite{Mitra2023}, operating in zero-shot mode without fine-tuning. 
We adapt CCoT for backlog generation in two steps: (i) extract a scene graph of UI components, attributes, and relationships; (ii) use this graph to generate backlog items (Listing~\ref{lst:ccot_prompt}).

\begin{tcolorbox}[customstyle]
    \begin{lstlisting}[style=customlisting, caption={CCoT prompt}, label={lst:ccot_prompt}]
Generate a scene graph for this group of mockups by identifying the objects (UI components) and their relationships. Summarize repeated elements and only provide unique attributes for each occurrence. Provide the output as a structured JSON object as shown below:
{
  "sceneGraph": {
    "objects": [],
    "relationships": {
      "ComponentType1": "relationship description"
    }
  }
}
Do not include any additional text or explanations outside the JSON object.
    \end{lstlisting}
\end{tcolorbox}

\noindent\textbf{Persona Prompt.} 
Personas are a well-established technique in requirements engineering for representing users and their needs\cite{Aoyama2005}. They help uncover overlooked requirements and improve communication with clients and team members. 
However, maintaining personas is resource-intensive, and teams often hesitate to keep them up to date~\cite{Wang2024}. 
Recent work has explored AI-driven persona generation to reduce the manual effort while maintaining a user-centred focus\cite{Kanij2023}.
We adopt Nguyen et al.’s persona template \cite{Khanh2017}, which incorporates storytelling elements to make user needs more actionable~\cite{Rainer2021}. 
Our two-step process first extracts personas from mockups using this template, then generates backlog items from the personas (Listing~\ref{lst:persona_prompt}).

\begin{tcolorbox}[customstyle]
\begin{lstlisting}[style=customlisting, caption={Persona prompt}, label={lst:persona_prompt}]
Consider the following group of mockups. Identify the potential user personas and roles using the provided JSON template:
{
  "Persona #i": {
    "who": { "name": "", "demographics": { "age": 0, "gender": "", "organization": "", "location": "" }, "role": "" },
    "what": { "requirement": "" },
    "why":  { "motivation": "" },
    "storytelling": { "story": "" }
  }
}
Provide only the JSON structure with no additional text
\end{lstlisting}
\end{tcolorbox}

\noindent\textbf{Architectural Context.} 
To examine whether basic architectural details affect task generation (RQ2), we run a simple ablation: the task prompt is executed with architectural context (C) ---a brief description of system attributes such as the technology stack---, and without it (NC), and we compare the outputs. Context descriptions were written individually for each project and were not standardized in length or level of detail. 
Since technology choices are not always evident from mockups, they can lead to variations in task decomposition. 
For example, a MERN (MongoDB, Express.js, React, Node.js) app may involve tasks such as defining MongoDB schemas, setting up Express API routes, or managing state in React, whereas a serverless backend may not (Listing~\ref{lst:task_prompt}).

\begin{tcolorbox}[customstyle]
    \begin{lstlisting}[style=customlisting, caption={Task prompt with architectural context}, label={lst:task_prompt}]
The following is the architectural context for the project: 

{Context description}

Generate the backlog items as specified:

**Tasks** 

- For each user story, list the tasks required to implement the feature. 
- Format: E{epic number}, US{user story number}, T{task number}: {Task}.

Provide only the specified items without additional text. 
    \end{lstlisting}
\end{tcolorbox}

\subsection{Data Collection}
\label{subsec:dataset}
We collected data from seven app development projects: three from AppComp, a software development company based in South America, and four from CapStone, a graduate project course with industrial  partners at a large university in Europe.
As Table~\ref{tab:datasets_summary} shows, the datasets covered various domains and consisted of app mockups and corresponding backlogs, extracted from issue tracking systems or project documentation. 
We unified the naming and structure to a predefined format for data processing, numbered the items and, where applicable, translated them into English. 

AppComp datasets (Projects 1–3) were collected from different teams working under company-level guidelines. 
Backlog items were organised hierarchically into epics, user stories, and tasks, and grouped with the corresponding mockups by app features. Each feature contained between two and seventeen mockups. 
The datasets varied in language and covered domains such as e-commerce, finance, and meal management.
CapStone datasets (Projects 4–7) were collected from student teams developing apps for industry or public sector clients. 
Backlog items included epics, scenarios, and tasks, but without a strict hierarchy. 
Each dataset contained a single set of 13–25 English-language mockups, covering domains such as social gathering, navigation, e-learning, and recipe management. Throughout the paper, we refer to user stories and scenarios collectively as US/S.   

\begin{table}[htbp]
\centering
\caption{Overview of datasets used in the study.}
\label{tab:datasets_summary}
\begin{threeparttable}
\resizebox{\columnwidth}{!}{
\begin{tabular}{@{}lllrrl@{}}
\toprule
\textbf{Dataset source} & \textbf{ID} & \textbf{Language} & \textbf{\# Mockups} & \textbf{\# Backlog items} & \textbf{App domain} \\
\midrule
AppComp & Project 1 & Portuguese & 2--17 & 109 & E-commerce \\
 & Project 2 & Spanish & 2--9 & 160 & Finance \\
 & Project 3 & Spanish & 4--9 & 132 & Meal management \\
\midrule
CapStone & Project 4 & English & 23 & 32 & Social gathering \\
 & Project 5 & English & 14 & 52 & Navigation \\
 & Project 6 & English & 13 & 64 & E-learning \\
 & Project 7 & English & 25 & 46 & Recipe management \\
\bottomrule
\end{tabular}
}
\begin{tablenotes}
\scriptsize
\item AppComp mockup counts indicate the range per feature group.
\end{tablenotes}
\end{threeparttable}
\end{table}

\subsection{Experimental Setup}
\subsubsection{Evaluation Metrics}
To evaluate the performance of each prompting strategy, we calculate precision, recall, and F1-scores as follows:

\begin{equation}
\text{Precision} = \frac{TP}{TP + FP}
\end{equation}

\begin{equation}
\text{Recall} = \frac{TP}{TP + FN}
\end{equation}

\begin{equation}
F1 = \frac{2 \times \text{Precision} \times \text{Recall}}
          {\text{Precision} + \text{Recall}}
\end{equation}

where true positive (TP) represents generated items  that match the ground truth backlog items, false positive (FP) represents items generated by the model that do not match any ground truth item, and false negative (FN) represents ground truth items that the model failed to generate.

\subsubsection{System Configuration and Prompt Execution}
\label{subsec:exp_setup}
We used OpenAI’s GPT-4o via its API with temperature set to zero for more reproducible outputs and the default token limit of 2048.
Each call included a system prompt (framing the model as an experienced requirements engineer focused on translating mockups into actionable backlog items) and a user message containing the prompt and Base64-encoded mockups. The same system prompt was used for all strategies (full text in the replication package).

Each backlog item (i.e.~epics, user stories, tasks) was generated separately using a dedicated prompt.
For the AppComp datasets, which followed a hierarchical structure, backlog items were generated sequentially: epics were produced first, followed by user stories, and finally tasks derived from those user stories. 
To maintain this hierarchy, the outputs generated in one stage were included as context in the subsequent prompt. For RQ2, even though we were interested in studying architectural context for tasks, the entire backlog was regenerated to preserve consistency. 
This led to small syntactic variations in other backlog items due to the non-deterministic nature of LLM responses.  
For CapStone datasets, which lacked a hierarchy, each item type was generated independently, without reference to previous outputs.

\subsubsection{Manual Matching of Generated Items}

After executing all three prompt types for each dataset, we compiled a spreadsheet that contained the ground‑truth backlog next to the outputs of the Baseline, CCoT, and Persona prompts. 
Given a ground‑truth (GT) item, we examined the outputs of each prompting strategy and looked for a candidate whose semantic intent matched the GT. 
Because each GT item is compared with many generated lines in the same column, this evaluation is a one‑to‑many retrieval task rather than a 50--50 binary classification problem. 
A candidate was accepted as a match when it (i) captured the core action/constraint of the GT, (ii) included all required elements, and (iii) matched the GT level of detail; minor wording or syntactic differences were ignored (e.g., GT: “When the user selects an event on the map, the system must display event details, including location, time, and a list of other users attending” vs. model: “When the user selects an event, the system must display event information and a list of attendees”). Conversely, items that were generic or only partially covered the GT were rejected. 
When several candidates satisfied the same GT, the most complete formulation was kept as the true positive (TP); the remainder were counted as false positives (FP) to avoid inflating recall. 
For the AppComp datasets this analysis was first performed inside each feature group and later aggregated. 

All backlog items in the seven datasets were peer-labelled independently by the first two authors. 
After the initial labelling, the authors reviewed each other’s labels to identify and resolve discrepancies. 
We checked the classifications for omissions or inconsistencies using ChatGPT-4o. Discrepancies from either the cross-review or ChatGPT verification were discussed and resolved by consensus.  

\subsubsection{Label Validity Check}\label{sec:design-sanity-check}
To assess reliability, we applied the GEval adequacy rubric~\cite{vongthongsri2025geval} (Claude~3 Sonnet, threshold~0.60) to all ground-truth/candidate pairs, reaching $\kappa = 0.60$ with the original labels.
An independent annotator re-evaluated a stratified 15\% sample ($n = 545$ pairs, balanced by item type, context variant, and dataset), yielding $\kappa = 0.87$ with the consensus labels, remaining high across context variants ($\kappa = 0.90$ / $0.84$) and dataset types ($\kappa = 0.85$ / $0.88$), validating the stability of the labelling criteria.
Full validity details are provided in the replication package.

\subsection{Perception of Development Teams}
\label{subsec:teams-perc}
We gathered feedback through face-to-face interviews with AppComp developers and CapStone teams from the studied projects.
A total of seven teams participated: three from AppComp (developers directly involved in backlog creation) and four from CapStone (team leads responsible for backlog management). 
The sessions lasted approximately 20 minutes and were conducted in a semi-structured format aimed at capturing practical perspectives on the generated backlog items. Each session began with four guiding questions:

\begin{enumerate}
    \item How would you use an AI tool for backlog generation/suggestion?
    \item When in the project life cycle would you use it?
    \item For which type of item would an AI support be most valuable?
    \item What challenges or potential issues do you see when using such a tool?
\end{enumerate}

Notes were taken during the sessions and later summarised by the authors, grouping responses according to the questions. The study involved no collection of personal or sensitive data. Participation was voluntary and verbal consent was obtained from all participants.
After answering the questions, participants reviewed a list of backlog items previously classified as FP in our evaluation. 
For each FP, participants were asked about the usefulness of the generated item for backlog creation and sprint planning: whether they would keep it as-is, keep it with changes, or discard it. These responses were aggregated to calculate a usefulness ratio for each prompting strategy, indicating the proportion of ``unexpected'' items considered beneficial to the teams.

\section{Results}
\label{sec:results}
We report on the results of our study along the three research questions: prompt strategy performance, impact of context, and perceived usefulness of backlog generation.

\subsection{Prompting Strategies (RQ1)}
\label{subsec:perf-analysis}

Table~\ref{tab:rq1-results} shows the precision, recall, and F1 scores for the Baseline, CCoT, and Persona prompting strategies across three types of backlog items: epics (E), user stories/scenarios (US/S), and tasks (T) for the AppComp and CapStone projects.
Overall, F1‑scores for tasks averaged 21–35\%, whereas performance for epics and user stories was substantially higher, with precision up to 50–80\% and recall up to 57–100\% across individual projects. 
The simple Baseline prompt performed best on average, with higher recall than precision, particularly for epics and user stories.

\begin{table*}[htbp]
\centering
\caption{Comparison of performance metrics for the three prompt strategies across all projects.}
\label{tab:rq1-results}
\resizebox{\textwidth}{!}{%
\begin{tabular}{@{}l|lllllllll|lllllllll|lllllllll@{}}
\cmidrule(l){2-28}
\multicolumn{1}{l}{} & \multicolumn{9}{c}{\textbf{Baseline}} & \multicolumn{9}{c}{\textbf{CCoT}} & \multicolumn{9}{c}{\textbf{Persona}} \\ 
\cmidrule(l){2-28}
\multicolumn{1}{l}{} & \multicolumn{3}{c}{Epics} & \multicolumn{3}{c}{US/S} & \multicolumn{3}{c|}{Tasks} & \multicolumn{3}{c}{Epics} & \multicolumn{3}{c}{US/S} & \multicolumn{3}{c|}{Tasks} & \multicolumn{3}{c}{Epics} & \multicolumn{3}{c}{US/S} & \multicolumn{3}{c}{Tasks} \\
\cmidrule(l){2-28}
\multicolumn{1}{l}{} 
& \multicolumn{1}{c}{P} & \multicolumn{1}{c}{R} & \multicolumn{1}{c|}{F1} 
& \multicolumn{1}{c}{P} & \multicolumn{1}{c}{R} & \multicolumn{1}{c|}{F1} 
& \multicolumn{1}{c}{P} & \multicolumn{1}{c}{R} & \multicolumn{1}{c|}{F1} 
& \multicolumn{1}{c}{P} & \multicolumn{1}{c}{R} & \multicolumn{1}{c|}{F1} 
& \multicolumn{1}{c}{P} & \multicolumn{1}{c}{R} & \multicolumn{1}{c|}{F1} 
& \multicolumn{1}{c}{P} & \multicolumn{1}{c}{R} & \multicolumn{1}{c|}{F1} 
& \multicolumn{1}{c}{P} & \multicolumn{1}{c}{R} & \multicolumn{1}{c|}{F1} 
& \multicolumn{1}{c}{P} & \multicolumn{1}{c}{R} & \multicolumn{1}{c|}{F1} 
& \multicolumn{1}{c}{P} & \multicolumn{1}{c}{R} & \multicolumn{1}{c}{F1} \\
\midrule
\multicolumn{1}{l|}{Project 1} 
& \cellcolor[HTML]{FAF1ED}50\% 
& \cellcolor[HTML]{D4E4F4}79\% 
& \multicolumn{1}{c|}{\cellcolor[HTML]{F1F6FC}61\%} 
& \cellcolor[HTML]{F6C8AA}34\% 
& \cellcolor[HTML]{F3F7FD}60\% 
& \multicolumn{1}{c|}{\cellcolor[HTML]{F8E0D1}43\%} 
& \cellcolor[HTML]{F3AD7D}21\% 
& \cellcolor[HTML]{FCFCFF}55\% 
& \cellcolor[HTML]{F5C5A4}31\% 
& \cellcolor[HTML]{F9E8DE}46\% 
& \cellcolor[HTML]{EBF2FB}65\% 
& \multicolumn{1}{c|}{\cellcolor[HTML]{FBFBFE}54\%} 
& \cellcolor[HTML]{F6CDB2}36\% 
& \cellcolor[HTML]{F3F7FD}60\% 
& \multicolumn{1}{c|}{\cellcolor[HTML]{F9E4D8}45\%} 
& \cellcolor[HTML]{F2A26B}21\% 
& \cellcolor[HTML]{F7D1B8}45\% 
& \cellcolor[HTML]{F3B185}29\% 
& \cellcolor[HTML]{FCFCFF}55\% 
& \cellcolor[HTML]{EEF4FC}63\% 
& \multicolumn{1}{c|}{\cellcolor[HTML]{F5F8FE}59\%} 
& \cellcolor[HTML]{F6C8AA}34\% 
& \cellcolor[HTML]{FAF1ED}50\% 
& \multicolumn{1}{c|}{\cellcolor[HTML]{F8D8C5}41\%} 
& \cellcolor[HTML]{F0904E}11\% 
& \cellcolor[HTML]{F4B991}28\% 
& \cellcolor[HTML]{F19C62}16\% \\
\multicolumn{1}{l|}{Project 2} 
& \cellcolor[HTML]{DAE8F6}75\% 
& \cellcolor[HTML]{BDD7EE}92\% 
& \multicolumn{1}{c|}{\cellcolor[HTML]{CDE1F3}83\%} 
& \cellcolor[HTML]{E6EFF9}68\% 
& \cellcolor[HTML]{D3E4F4}79\% 
& \multicolumn{1}{c|}{\cellcolor[HTML]{DDEAF7}73\%} 
& \cellcolor[HTML]{F7CFB5}37\% 
& \cellcolor[HTML]{F7F9FE}58\% 
& \cellcolor[HTML]{F9E4D7}45\% 
& \cellcolor[HTML]{EAF1FA}66\% 
& \cellcolor[HTML]{D1E3F4}81\% 
& \multicolumn{1}{c|}{\cellcolor[HTML]{DEEBF7}72\%} 
& \cellcolor[HTML]{FAF1ED}50\% 
& \cellcolor[HTML]{E9F1FA}66\% 
& \multicolumn{1}{c|}{\cellcolor[HTML]{F8FAFE}57\%} 
& \cellcolor[HTML]{F4B58C}26\% 
& \cellcolor[HTML]{F8D9C6}38\% 
& \cellcolor[HTML]{F5C4A3}31\% 
& \cellcolor[HTML]{ECF3FB}64\% 
& \cellcolor[HTML]{F0F5FC}62\% 
& \multicolumn{1}{c|}{\cellcolor[HTML]{EEF4FC}63\%} 
& \cellcolor[HTML]{F4F7FD}60\% 
& \cellcolor[HTML]{EFF5FC}62\% 
& \multicolumn{1}{c|}{\cellcolor[HTML]{F2F6FD}61\%} 
& \cellcolor[HTML]{F3B387}26\% 
& \cellcolor[HTML]{F8DDCD}43\% 
& \cellcolor[HTML]{F5C3A2}32\% \\
\multicolumn{1}{l|}{Project 3} 
& \cellcolor[HTML]{F8FAFE}57\% 
& \cellcolor[HTML]{CCE0F3}83\% 
& \multicolumn{1}{c|}{\cellcolor[HTML]{E6EFF9}68\%} 
& \cellcolor[HTML]{F9E6DC}46\% 
& \cellcolor[HTML]{D2E3F4}80\% 
& \multicolumn{1}{c|}{\cellcolor[HTML]{F6F9FE}58\%} 
& \cellcolor[HTML]{F2A36E}20\% 
& \cellcolor[HTML]{FAEDE7}49\% 
& \cellcolor[HTML]{F4B991}28\% 
& \cellcolor[HTML]{F6F9FE}58\% 
& \cellcolor[HTML]{BFD8EF}91\% 
& \multicolumn{1}{c|}{\cellcolor[HTML]{E0ECF8}71\%} 
& \cellcolor[HTML]{F9E3D6}45\% 
& \cellcolor[HTML]{E3EEF9}69\% 
& \multicolumn{1}{c|}{\cellcolor[HTML]{FCFCFF}54\%} 
& \cellcolor[HTML]{F2A774}21\% 
& \cellcolor[HTML]{F8DBC9}41\% 
& \cellcolor[HTML]{F4B891}28\% 
& \cellcolor[HTML]{FBFAFC}54\% 
& \cellcolor[HTML]{EFF4FC}63\% 
& \multicolumn{1}{c|}{\cellcolor[HTML]{F7F9FE}58\%} 
& \cellcolor[HTML]{F9E9DF}47\% 
& \cellcolor[HTML]{F4F7FD}59\% 
& \multicolumn{1}{c|}{\cellcolor[HTML]{FBF7F7}52\%} 
& \cellcolor[HTML]{F3B287}26\% 
& \cellcolor[HTML]{FCFCFF}54\% 
& \cellcolor[HTML]{F6CAAD}35\% \\
\multicolumn{1}{l|}{\textbf{Avg.}} 
& \cellcolor[HTML]{F2F6FD}\textbf{61\%} 
& \cellcolor[HTML]{CADFF2}\textbf{85\%} 
& \multicolumn{1}{c|}{\cellcolor[HTML]{E1EDF8}\textbf{71\%}} 
& \cellcolor[HTML]{FAEFEA}\textbf{49\%} 
& \cellcolor[HTML]{DDEAF7}\textbf{73\%} 
& \multicolumn{1}{c|}{\cellcolor[HTML]{F6F9FE}\textbf{58\%}} 
& \cellcolor[HTML]{F4B58B}\textbf{26\%} 
& \cellcolor[HTML]{FBFAFC}\textbf{54\%} 
& \cellcolor[HTML]{F6CBAF}\textbf{35\%} 
& \cellcolor[HTML]{F8FAFE}\textbf{57\%} 
& \cellcolor[HTML]{D3E4F4}\textbf{79\%} 
& \multicolumn{1}{c|}{\cellcolor[HTML]{E9F1FA}\textbf{66\%}} 
& \cellcolor[HTML]{F8E0D2}\textbf{44\%} 
& \cellcolor[HTML]{EAF2FB}\textbf{65\%} 
& \multicolumn{1}{c|}{\cellcolor[HTML]{FBF6F5}\textbf{52\%}} 
& \cellcolor[HTML]{F2AA79}\textbf{23\%} 
& \cellcolor[HTML]{F7D7C3}\textbf{42\%} 
& \cellcolor[HTML]{F4BA93}\textbf{29\%} 
& \cellcolor[HTML]{F7F9FE}\textbf{57\%} 
& \cellcolor[HTML]{EFF5FC}\textbf{62\%} 
& \multicolumn{1}{c|}{\cellcolor[HTML]{F4F7FD}\textbf{60\%}} 
& \cellcolor[HTML]{F9E9DF}\textbf{47\%} 
& \cellcolor[HTML]{F8FAFE}\textbf{57\%} 
& \multicolumn{1}{c|}{\cellcolor[HTML]{FBF4F2}\textbf{51\%}} 
& \cellcolor[HTML]{F2A774}\textbf{21\%} 
& \cellcolor[HTML]{F8DBCA}\textbf{42\%} 
& \cellcolor[HTML]{F4B890}\textbf{28\%} \\
\midrule
\multicolumn{1}{l|}{Project 4} 
& \cellcolor[HTML]{E8F0FA}63\% 
& \cellcolor[HTML]{BDD7EE}100\% 
& \multicolumn{1}{c|}{\cellcolor[HTML]{D8E7F6}77\%} 
& \cellcolor[HTML]{FCFCFF}44\% 
& \cellcolor[HTML]{D4E5F5}80\% 
& \multicolumn{1}{c|}{\cellcolor[HTML]{EEF4FC}57\%} 
& \cellcolor[HTML]{FAF2EF}41\% 
& \cellcolor[HTML]{F8D8C4}32\% 
& \cellcolor[HTML]{F9E3D7}36\% 
& \cellcolor[HTML]{E8F0FA}63\% 
& \cellcolor[HTML]{BDD7EE}100\% 
& \multicolumn{1}{c|}{\cellcolor[HTML]{D8E7F6}77\%} 
& \cellcolor[HTML]{F6F9FE}50\% 
& \cellcolor[HTML]{EBF2FB}60\% 
& \multicolumn{1}{c|}{\cellcolor[HTML]{F1F6FC}55\%} 
& \cellcolor[HTML]{FAECE5}39\% 
& \cellcolor[HTML]{F8D8C4}32\% 
& \cellcolor[HTML]{F9E1D3}35\% 
& \cellcolor[HTML]{E3EEF9}67\% 
& \cellcolor[HTML]{D4E5F5}80\% 
& \multicolumn{1}{c|}{\cellcolor[HTML]{DCEAF7}73\%} 
& \cellcolor[HTML]{F6F9FE}50\% 
& \cellcolor[HTML]{EBF2FB}60\% 
& \multicolumn{1}{c|}{\cellcolor[HTML]{F1F6FC}55\%} 
& \cellcolor[HTML]{F8DCCB}33\% 
& \cellcolor[HTML]{F5BE9A}23\% 
& \cellcolor[HTML]{F6CAAE}27\% \\
\multicolumn{1}{l|}{Project 5} 
& \cellcolor[HTML]{EEF4FC}57\% 
& \cellcolor[HTML]{F6F9FE}50\% 
& \multicolumn{1}{c|}{\cellcolor[HTML]{F2F7FD}53\%} 
& \cellcolor[HTML]{F6F9FE}50\% 
& \cellcolor[HTML]{BDD7EE}100\% 
& \multicolumn{1}{c|}{\cellcolor[HTML]{E3EEF9}67\%} 
& \cellcolor[HTML]{FAFBFF}47\% 
& \cellcolor[HTML]{F3AE80}17\% 
& \cellcolor[HTML]{F5C5A4}25\% 
& \cellcolor[HTML]{D4E5F5}80\% 
& \cellcolor[HTML]{EEF4FC}57\% 
& \multicolumn{1}{c|}{\cellcolor[HTML]{E3EEF9}67\%} 
& \cellcolor[HTML]{FAEFEA}40\% 
& \cellcolor[HTML]{E3EEF9}67\% 
& \multicolumn{1}{c|}{\cellcolor[HTML]{F6F9FE}50\%} 
& \cellcolor[HTML]{F5BE9A}23\% 
& \cellcolor[HTML]{F1A069}12\% 
& \cellcolor[HTML]{F3AB7A}16\% 
& \cellcolor[HTML]{EBF2FB}60\% 
& \cellcolor[HTML]{F9E8DE}38\% 
& \multicolumn{1}{c|}{\cellcolor[HTML]{FBFBFF}46\%} 
& \cellcolor[HTML]{FAEFEA}40\% 
& \cellcolor[HTML]{E3EEF9}67\% 
& \multicolumn{1}{c|}{\cellcolor[HTML]{F6F9FE}50\%} 
& \cellcolor[HTML]{EDF3FB}58\% 
& \cellcolor[HTML]{F3AE80}17\% 
& \cellcolor[HTML]{F6C9AB}26\% \\
\multicolumn{1}{l|}{Project 6} 
& \cellcolor[HTML]{F7CFB5}29\% 
& \cellcolor[HTML]{F4B68D}20\% 
& \multicolumn{1}{c|}{\cellcolor[HTML]{F5C09E}24\%} 
& \cellcolor[HTML]{F8DCCB}33\% 
& \cellcolor[HTML]{F6F9FE}50\% 
& \multicolumn{1}{c|}{\cellcolor[HTML]{FAEFEA}40\%} 
& \cellcolor[HTML]{F4BA94}21\% 
& \cellcolor[HTML]{F0904E}6\% 
& \cellcolor[HTML]{F1995D}10\% 
& \cellcolor[HTML]{FBF7F7}43\% 
& \cellcolor[HTML]{F7D3BC}30\% 
& \multicolumn{1}{c|}{\cellcolor[HTML]{F9E2D4}35\%} 
& \cellcolor[HTML]{F9E8DE}38\% 
& \cellcolor[HTML]{F6F9FE}50\% 
& \multicolumn{1}{c|}{\cellcolor[HTML]{FBF7F7}43\%} 
& \cellcolor[HTML]{F4BA94}21\% 
& \cellcolor[HTML]{F0904E}6\% 
& \cellcolor[HTML]{F1995D}10\% 
& \cellcolor[HTML]{D4E5F5}80\% 
& \cellcolor[HTML]{FAEFEA}40\% 
& \multicolumn{1}{c|}{\cellcolor[HTML]{F2F7FD}53\%} 
& \cellcolor[HTML]{F8DCCB}33\% 
& \cellcolor[HTML]{F6F9FE}50\% 
& \multicolumn{1}{c|}{\cellcolor[HTML]{FAEFEA}40\%} 
& \cellcolor[HTML]{F7D3BC}30\% 
& \cellcolor[HTML]{F0904E}6\% 
& \cellcolor[HTML]{F19B60}10\% \\
\multicolumn{1}{l|}{Project 7} 
& \cellcolor[HTML]{FAEFEA}40\% 
& \cellcolor[HTML]{E3EEF9}67\% 
& \multicolumn{1}{c|}{\cellcolor[HTML]{F6F9FE}50\%} 
& \cellcolor[HTML]{FAEFEA}40\% 
& \cellcolor[HTML]{BDD7EE}100\% 
& \multicolumn{1}{c|}{\cellcolor[HTML]{EEF4FC}57\%} 
& \cellcolor[HTML]{F7CFB5}29\% 
& \cellcolor[HTML]{F19B60}10\% 
& \cellcolor[HTML]{F2A976}15\% 
& \cellcolor[HTML]{EBF2FB}60\% 
& \cellcolor[HTML]{BDD7EE}100\% 
& \multicolumn{1}{c|}{\cellcolor[HTML]{DAE8F6}75\%} 
& \cellcolor[HTML]{F6F9FE}50\% 
& \cellcolor[HTML]{BDD7EE}100\% 
& \multicolumn{1}{c|}{\cellcolor[HTML]{E3EEF9}67\%} 
& \cellcolor[HTML]{F6C9AC}27\% 
& \cellcolor[HTML]{F19B60}10\% 
& \cellcolor[HTML]{F2A875}15\% 
& \cellcolor[HTML]{FAEFEA}40\% 
& \cellcolor[HTML]{E3EEF9}67\% 
& \multicolumn{1}{c|}{\cellcolor[HTML]{F6F9FE}50\%} 
& \cellcolor[HTML]{F8DCCB}33\% 
& \cellcolor[HTML]{E3EEF9}67\% 
& \multicolumn{1}{c|}{\cellcolor[HTML]{FCFCFF}44\%} 
& \cellcolor[HTML]{FAEFEA}40\% 
& \cellcolor[HTML]{F19B60}10\% 
& \cellcolor[HTML]{F3AC7C}16\% \\
\multicolumn{1}{l|}{\textbf{Avg.}} 
& \cellcolor[HTML]{FAFBFF}\textbf{47\%} 
& \cellcolor[HTML]{ECF3FB}\textbf{59\%} 
& \multicolumn{1}{c|}{\cellcolor[HTML]{F5F8FE}\textbf{51\%}} 
& \cellcolor[HTML]{FBF4F3}\textbf{42\%} 
& \cellcolor[HTML]{D1E3F4}\textbf{83\%} 
& \multicolumn{1}{c|}{\cellcolor[HTML]{F0F5FC}\textbf{55\%}} 
& \cellcolor[HTML]{F8DFD0}\textbf{34\%} 
& \cellcolor[HTML]{F3AC7C}\textbf{16\%} 
& \cellcolor[HTML]{F4BA94}\textbf{21\%} 
& \cellcolor[HTML]{E9F1FA}\textbf{61\%} 
& \cellcolor[HTML]{DDEAF7}\textbf{72\%} 
& \multicolumn{1}{c|}{\cellcolor[HTML]{E7F0FA}\textbf{63\%}} 
& \cellcolor[HTML]{FBFBFE}\textbf{44\%} 
& \cellcolor[HTML]{E0ECF8}\textbf{69\%} 
& \multicolumn{1}{c|}{\cellcolor[HTML]{F2F6FD}\textbf{54\%}} 
& \cellcolor[HTML]{F6CBB0}\textbf{27\%} 
& \cellcolor[HTML]{F2A977}\textbf{15\%} 
& \cellcolor[HTML]{F3B388}\textbf{19\%} 
& \cellcolor[HTML]{E9F1FA}\textbf{62\%} 
& \cellcolor[HTML]{EFF5FC}\textbf{56\%} 
& \multicolumn{1}{c|}{\cellcolor[HTML]{F0F5FC}\textbf{56\%}} 
& \cellcolor[HTML]{FAEDE6}\textbf{39\%} 
& \cellcolor[HTML]{EAF2FA}\textbf{61\%} 
& \multicolumn{1}{c|}{\cellcolor[HTML]{F9FBFF}\textbf{47\%}} 
& \cellcolor[HTML]{FAF0EC}\textbf{40\%} 
& \cellcolor[HTML]{F2A672}\textbf{14\%} 
& \cellcolor[HTML]{F4B68D}\textbf{20\%} \\
\bottomrule
\end{tabular}%
}
\end{table*}

In AppComp datasets, where the mockups  were overall simpler and had a lower fidelity to the final UI, the Baseline prompt yields strong results for epics, achieving an average of 85\% recall with 61\% precision (71\% F1). 
Similarly, for user stories, Baseline attains 73\% average recall with precision below 50\% (yielding an F1 of 58\%). 
In contrast, CapStone datasets show Baseline epics at 59\% recall and 47\% precision (51\% F1), while scenarios reach 83\% recall (42\% precision, 55\% F1). 
Thus, Baseline tends to produce high recall (especially for higher-level backlog items) even if precision is compromised.
This is most evident with tasks, where Baseline achieves the highest F1 scores, while Persona sometimes reaches higher recall but at lower precision.

CCoT prompts maintain high recall for epics (79\% average for AppComp and 72\% for CapStone) and precision fluctuates between 57\% and 61\%, resulting in F1 scores in the mid-60s range. 
For user stories, recall spans from 65\% (52\% F1) in the AppComp datasets to 69\% (54\% F1) in CapStone datasets, with similar precision across both contexts. 

The Persona prompts show mixed results when compared to both Baseline and CCoT. 
In AppComp projects, recall for epics (62\%) falls below CCoT (79\%) and the Baseline (85\%), although its precision (57\%) is comparable to CCoT (57\%). 
In CapStone projects, Persona’s recall for epics also remains with 56\% lower than CCoT (72\%) and Baseline (59\%), but its precision (62\%) surpasses both. 

Generating the tasks remains challenging for all prompts. 
For CCoT, recall in the AppComp dataset is 42\% and drops to 15\% in the CapStone dataset. 
Although precision rises from 23\% to 27\%, the low recall in the CapStone setting produces a 19\% F1. 
Baseline also struggles with CapStone tasks, reaching only 16\% recall, while Persona shows 42\% recall for tasks in the AppComp dataset and 14\% in the CapStone dataset. 
In every case, F1 scores stay rather low, indicating that none of the studied prompts consistently generates an accurate set of tasks only based on the mockups.

One possible explanation for these differences is that the model only had access to mockups (zero-shot setting), so its ability to generate accurate backlog items depends on how explicitly each functionality and domain-specific processes were depicted visually. 
For instance, in Project 2, the UI presented elements like address management forms, notifications, and financial transactions, with clear visuals for adding, editing, and deleting entries. 
The model could generate backlog items aligned with the ground truth, contributing to high recall for epics (92\%) and user stories (79\%).
By contrast, in Project 6, the UI was more abstract, focusing on guided learning and AI-assisted decision-making. 
The mockups showed progress tracking, a chat feature, and interactive modules, but functions such as automated planning or expert consultation were implied rather than shown. 
Tasks related to AI-driven suggestions were often missing or too broadly defined.

Overall, aggregated across projects and item types, Persona achieves the highest precision (44\%), whereas Baseline attains the best recall (62\%) and F1 (49\%), indicating that the added complexity of CCoT and Persona did not consistently improve over the zero-shot baseline.
However, the results show a precision–recall trade-off across prompts. 
Baseline yields broader coverage but lower precision, especially for tasks. 
CCoT and Persona improve precision in some cases but miss more items. 
No single prompt works best across all projects: structured reasoning (CCoT) is more effective for higher-level items such as epics and user stories, whereas direct prompting (Baseline) works better for detailed tasks.
Persona can be useful for reframing items from a user perspective, highlighting motivations and interactions not explicit in mockups.
For instance, in Project 5, the ground truth described a scenario where the user "traveller" adds a new vehicle profile, selecting motorisation type (e.g., gas/electric) and saving the details for future trips, while the Persona output reframes this as an `eco-conscious driver'' considering fuel efficiency and environmental impact.
Combining prompting strategies may therefore be necessary for accurate backlog generation across levels.

\subsection{Impact of Architectural Context (RQ2)}
\label{subsec:context}

To assess the impact of architectural context, we focused on task generation, as tasks are most likely to depend on system internals.
We studied this in three steps: (1) compare performance with and without context across prompts; (2) analyse performance by task type; and (3) inspect examples to characterize changes in LLM behaviour. Table~\ref{tab:arch_variation} reports precision, recall, and F1 for each prompt under two conditions: with (C) and without (NC) architectural context across the seven datasets, together with the difference ({$\Delta$}). Positive {$\Delta$} indicates improved performance with context.

\begin{table*}[]
\centering
\caption{Impact of adding architectural context to predict backlog tasks.}
\label{tab:arch_variation}
\resizebox{\textwidth}{!}{%
\begin{tabular}{@{}crrrrrrrrrrrrrrrrrrrrrrrrrrr@{}}
\toprule
 & \multicolumn{9}{c}{Baseline} & \multicolumn{9}{c}{CCoT} & \multicolumn{9}{c}{Persona} \\ \midrule
\multicolumn{1}{c|}{} & \multicolumn{3}{c}{Precision} & \multicolumn{3}{c}{Recall} & \multicolumn{3}{c|}{F1} & \multicolumn{3}{c}{Precision} & \multicolumn{3}{c}{Recall} & \multicolumn{3}{c|}{F1} & \multicolumn{3}{c}{Precision} & \multicolumn{3}{c}{Recall} & \multicolumn{3}{c}{F1} \\ \midrule
\multicolumn{1}{c|}{} & \multicolumn{1}{c}{NC} & \multicolumn{1}{c}{C} & \multicolumn{1}{c|}{$\Delta$} & \multicolumn{1}{c}{NC} & \multicolumn{1}{c}{C} & \multicolumn{1}{c|}{$\Delta$} & \multicolumn{1}{c}{NC} & \multicolumn{1}{c}{C} & \multicolumn{1}{c|}{$\Delta$} & \multicolumn{1}{c}{NC} & \multicolumn{1}{c}{C} & \multicolumn{1}{c|}{$\Delta$} & \multicolumn{1}{c}{NC} & \multicolumn{1}{c}{C} & \multicolumn{1}{c|}{$\Delta$} & \multicolumn{1}{c}{NC} & \multicolumn{1}{c}{C} & \multicolumn{1}{c|}{$\Delta$} & \multicolumn{1}{c}{NC} & \multicolumn{1}{c}{C} & \multicolumn{1}{c|}{$\Delta$} & \multicolumn{1}{c}{NC} & \multicolumn{1}{c}{C} & \multicolumn{1}{c|}{$\Delta$} & \multicolumn{1}{c}{NC} & \multicolumn{1}{c}{C} & \multicolumn{1}{c}{$\Delta$} \\ \midrule
\multicolumn{1}{c|}{Project 1} & 21\% & 16\% & \multicolumn{1}{r|}{{\color[HTML]{9C0006} -5\%}} & 55\% & 47\% & \multicolumn{1}{r|}{{\color[HTML]{9C0006} -8\%}} & 31\% & 24\% & \multicolumn{1}{r|}{{\color[HTML]{9C0006} -7\%}} & 21\% & 24\% & \multicolumn{1}{r|}{{\color[HTML]{70AD47} 3\%}} & 45\% & 47\% & \multicolumn{1}{r|}{{\color[HTML]{70AD47} 2\%}} & 39\% & 32\% & \multicolumn{1}{r|}{{\color[HTML]{9C0006} -7\%}} & 11\% & 9\% & \multicolumn{1}{r|}{{\color[HTML]{9C0006} -2\%}} & 28\% & 23\% & \multicolumn{1}{r|}{{\color[HTML]{9C0006} -5\%}} & 16\% & 13\% & {\color[HTML]{9C0006} -3\%} \\
\multicolumn{1}{c|}{Project 2} & 37\% & 37\% & \multicolumn{1}{r|}{{ 0\%}} & 58\% & 61\% & \multicolumn{1}{r|}{{\color[HTML]{70AD47} 3\%}} & 45\% & 46\% & \multicolumn{1}{r|}{{\color[HTML]{70AD47} 1\%}} & 26\% & 35\% & \multicolumn{1}{r|}{{\color[HTML]{70AD47} 9\%}} & 38\% & 53\% & \multicolumn{1}{r|}{{\color[HTML]{70AD47} 15\%}} & 31\% & 42\% & \multicolumn{1}{r|}{{\color[HTML]{70AD47} 11\%}} & 26\% & 29\% & \multicolumn{1}{r|}{{\color[HTML]{70AD47} 3\%}} & 43\% & 46\% & \multicolumn{1}{r|}{{\color[HTML]{70AD47} 3\%}} & 32\% & 35\% & {\color[HTML]{70AD47} 3\%} \\
\multicolumn{1}{c|}{Project 3} & 20\% & 22\% & \multicolumn{1}{r|}{{\color[HTML]{70AD47} 2\%}} & 49\% & 54\% & \multicolumn{1}{r|}{{\color[HTML]{70AD47} 6\%}} & 28\% & 31\% & \multicolumn{1}{r|}{{\color[HTML]{70AD47} 3\%}} & 21\% & 21\% & \multicolumn{1}{r|}{{ 0\%}} & 41\% & 44\% & \multicolumn{1}{r|}{{\color[HTML]{70AD47} 3\%}} & 28\% & 28\% & \multicolumn{1}{r|}{{ 0\%}} & 26\% & 20\% & \multicolumn{1}{r|}{{\color[HTML]{9C0006} -5\%}} & 56\% & 46\% & \multicolumn{1}{r|}{{\color[HTML]{9C0006} -8\%}} & 35\% & 28\% & {\color[HTML]{9C0006} -7\%} \\
\multicolumn{1}{l|}{\textbf{Avg.}} & \textbf{26\%} & \textbf{25\%} & \multicolumn{1}{r|}{{\color[HTML]{9C0006} \textbf{-1\%}}} & \textbf{54\%} & \textbf{54\%} & \multicolumn{1}{r|}{\textbf{0\%}} & \textbf{35\%} & \textbf{34\%} & \multicolumn{1}{r|}{{\color[HTML]{9C0006} \textbf{-1\%}}} & \textbf{23\%} & \textbf{27\%} & \multicolumn{1}{r|}{{\color[HTML]{70AD47} \textbf{4\%}}} & \textbf{41\%} & \textbf{48\%} & \multicolumn{1}{r|}{{\color[HTML]{70AD47} \textbf{7\%}}} & \textbf{33\%} & \textbf{34\%} & \multicolumn{1}{r|}{{\color[HTML]{70AD47} \textbf{1\%}}} & \textbf{21\%} & \textbf{19\%} & \multicolumn{1}{r|}{{\color[HTML]{9C0006} \textbf{-1\%}}} & \textbf{42\%} & \textbf{38\%} & \multicolumn{1}{r|}{{\color[HTML]{9C0006} \textbf{-3\%}}} & \textbf{28\%} & \textbf{25\%} & {\color[HTML]{9C0006} \textbf{-2\%}} \\ \midrule
\multicolumn{1}{c|}{Project 4} & 41\% & 65\% & \multicolumn{1}{r|}{{\color[HTML]{70AD47} 24\%}} & 32\% & 50\% & \multicolumn{1}{r|}{{\color[HTML]{70AD47} 18\%}} & 36\% & 56\% & \multicolumn{1}{r|}{{\color[HTML]{70AD47} 21\%}} & 39\% & 47\% & \multicolumn{1}{r|}{{\color[HTML]{70AD47} 8\%}} & 32\% & 36\% & \multicolumn{1}{r|}{{\color[HTML]{70AD47} 5\%}} & 35\% & 41\% & \multicolumn{1}{r|}{{\color[HTML]{70AD47} 6\%}} & 33\% & 60\% & \multicolumn{1}{r|}{{\color[HTML]{70AD47} 27\%}} & 23\% & 41\% & \multicolumn{1}{r|}{{\color[HTML]{70AD47} 18\%}} & 27\% & 49\% & {\color[HTML]{70AD47} 22\%} \\
\multicolumn{1}{c|}{Project 5} & 47\% & 53\% & \multicolumn{1}{r|}{{\color[HTML]{70AD47} 6\%}} & 17\% & 24\% & \multicolumn{1}{r|}{{\color[HTML]{70AD47} 7\%}} & 25\% & 33\% & \multicolumn{1}{r|}{{\color[HTML]{70AD47} 8\%}} & 23\% & 32\% & \multicolumn{1}{r|}{{\color[HTML]{70AD47} 9\%}} & 12\% & 18\% & \multicolumn{1}{r|}{{\color[HTML]{70AD47} 5\%}} & 16\% & 23\% & \multicolumn{1}{r|}{{\color[HTML]{70AD47} 7\%}} & 58\% & 67\% & \multicolumn{1}{r|}{{\color[HTML]{70AD47} 8\%}} & 17\% & 20\% & \multicolumn{1}{r|}{{\color[HTML]{70AD47} 2\%}} & 26\% & 30\% & {\color[HTML]{70AD47} 4\%} \\
\multicolumn{1}{c|}{Project 6} & 21\% & 47\% & \multicolumn{1}{r|}{{\color[HTML]{70AD47} 25\%}} & 6\% & 15\% & \multicolumn{1}{r|}{{\color[HTML]{70AD47} 8\%}} & 10\% & 22\% & \multicolumn{1}{r|}{{\color[HTML]{70AD47} 13\%}} & 21\% & 50\% & \multicolumn{1}{r|}{{\color[HTML]{70AD47} 29\%}} & 6\% & 13\% & \multicolumn{1}{r|}{{\color[HTML]{70AD47} 6\%}} & 10\% & 20\% & \multicolumn{1}{r|}{{\color[HTML]{70AD47} 10\%}} & 30\% & 44\% & \multicolumn{1}{r|}{{\color[HTML]{70AD47} 14\%}} & 6\% & 8\% & \multicolumn{1}{r|}{{\color[HTML]{70AD47} 2\%}} & 10\% & 14\% & {\color[HTML]{70AD47} 4\%} \\
\multicolumn{1}{c|}{Project 7} & 29\% & 58\% & \multicolumn{1}{r|}{{\color[HTML]{70AD47} 30\%}} & 10\% & 18\% & \multicolumn{1}{r|}{{\color[HTML]{70AD47} 8\%}} & 15\% & 27\% & \multicolumn{1}{r|}{{\color[HTML]{70AD47} 12\%}} & 27\% & 38\% & \multicolumn{1}{r|}{{\color[HTML]{70AD47} 12\%}} & 10\% & 13\% & \multicolumn{1}{r|}{{\color[HTML]{70AD47} 3\%}} & 15\% & 19\% & \multicolumn{1}{r|}{{\color[HTML]{70AD47} 4\%}} & 40\% & 75\% & \multicolumn{1}{r|}{{\color[HTML]{70AD47} 35\%}} & 10\% & 23\% & \multicolumn{1}{r|}{{\color[HTML]{70AD47} 13\%}} & 16\% & 35\% & {\color[HTML]{70AD47} 19\%} \\
\multicolumn{1}{l|}{\textbf{Avg.}} & \textbf{34\%} & \textbf{56\%} & \multicolumn{1}{r|}{{\color[HTML]{70AD47} \textbf{21\%}}} & \textbf{16\%} & \textbf{27\%} & \multicolumn{1}{r|}{{\color[HTML]{70AD47} \textbf{10\%}}} & \textbf{21\%} & \textbf{35\%} & \multicolumn{1}{r|}{{\color[HTML]{70AD47} \textbf{13\%}}} & \textbf{27\%} & \textbf{42\%} & \multicolumn{1}{r|}{{\color[HTML]{70AD47} \textbf{14\%}}} & \textbf{15\%} & \textbf{20\%} & \multicolumn{1}{r|}{{\color[HTML]{70AD47} \textbf{5\%}}} & \textbf{19\%} & \textbf{26\%} & \multicolumn{1}{r|}{{\color[HTML]{70AD47} \textbf{7\%}}} & \textbf{40\%} & \textbf{62\%} & \multicolumn{1}{r|}{{\color[HTML]{70AD47} \textbf{21\%}}} & \textbf{14\%} & \textbf{23\%} & \multicolumn{1}{r|}{{\color[HTML]{70AD47} \textbf{9\%}}} & \textbf{20\%} & \textbf{32\%} & {\color[HTML]{70AD47} \textbf{12\%}} \\ \bottomrule
\end{tabular}%
}
\end{table*}

CCoT benefited most from architectural context: F1 improved in five of seven projects, stayed unchanged in one, and decreased in one, for an average gain of 4\% (up to +11\%). In contrast, the Persona prompt decreased in Projects~1 ($\Delta=-3\%$) and~3 ($\Delta=-7\%$), and Baseline dropped in Project~1 ($\Delta=-7\%$), suggesting that added system-level information can sometimes shift the model away from the intended abstraction level. CCoT remained stable or improved across all datasets, with larger gains in Projects~2 (+11\%), 5 (+7\%), and 6 (+10\%), and smaller gains in Projects~4 (+6\%) and 7 (+4\%).

The declines are consistent with the ground-truth task phrasing in Projects~1 and~3.
In Project~1, GT tasks were high-level (e.g., ``implement the UI to show product image, name, and price''), emphasising functionality over implementation; context may therefore have introduced unnecessary detail without improving alignment.
In Project~3, context shifted outputs toward low-level backend operations (e.g., database or service integration) rather than functional goals. For example, the GT task ``Integrate weather data API to fetch daily forecasts'' matched in the NC run but not in the C run, where the model prioritised UI-to-backend connections instead.

\textbf{Impact by Task Type.} To identify which kinds of tasks benefit most from architectural context, we categorized each task into high-level types such as  Backend, UI, Logic and Messaging and computed recall separately for each. 
This grouping was performed iteratively by clustering similar task descriptions using keyword-based heuristics and regular expressions (e.g., “endpoint”, “API”, “controller” for  Backend; “button”, “screen”, “click” for UI). 

The most notable improvements occurred in the Backend category, where recall increased from 34\% without context to 52\% with context. 
This improvement was statistically significant (McNemar's exact test, $p < 0.001$). 
Other categories, such as Integration and ModuleFeature, also showed large recall gains (over +13\%), although their sample sizes were smaller and statistical significance was marginal ($p = 0.062$). The Data and Messaging categories showed smaller gains (both +4\%), and in both cases, the differences were not statistically significant.

These results suggest that simple architectural context helps most for implementation-oriented tasks involving backend services, integration layers, and technical modules. By contrast, more abstract categories such as Messaging or Data showed limited improvement, likely because they rely on domain knowledge not captured by system-level architecture.

\textbf{Qualitative Analysis of Task Improvements.} We compared model outputs with and without context to characterise how architectural context changes task generation. 
We examined cases where (a) context enabled a correct task previously missed, (b) context improved specificity, and (c) context degraded an otherwise correct output.

Candidate examples were drawn from outputs where human annotation and LLM evaluation (via GEval introduced in Section \ref{sec:design-sanity-check}) showed differences between the context and no-context conditions. For each selected case, we examined what changed in the wording of the generated task and how these changes aligned with the ground truth. On average, GEval scores increased most for Integration tasks (+18\%) and Backend tasks (+14\%), with smaller but still positive gains for Messaging (+8\%) and Data (+6\%). Improvements were minimal for ModuleFeature tasks (+3\%). These values reflect changes in semantic adequacy when architectural context is added.

GEval scores align with the recall analysis: gains are clearest for system-level categories (Backend, Integration), while abstract or UI-oriented tasks show mild degradation when context is added. The following examples illustrate these effects.

\textbf{(1) Context improves specificity:}
\begin{answerbox}
\textbf{Project 2, Prompt = Baseline}\\

\textbf{GT:} "Implement functionality for users to upload ID document photos with file format and size restrictions." \\[4pt]
\textbf{NC:} "Implement photo upload functionality." \\
\textit{GEval = 50\%} \\[4pt]
\textbf{C:} Implement the photo upload functionality for document verification. \\
\textit{GEval = 70\%, $\Delta$ = 20\%} \\[4pt]
\textbf{What changed:} NC is generic; with context, the upload is framed as document verification, improving GT alignment by increasing semantic specificity.
\end{answerbox}

\textbf{(2) Context reduces specificity:}

\begin{answerbox}
\textbf{Project 1, Prompt = CCoT}\\

\textbf{GT:} "Develop logic to sort products based on selected criteria." \\[4pt]
\textbf{NC:} "Update product list based on selected sorting option." \\
\textit{GEval = 80\%} \\[4pt]
\textbf{C:} "Create API endpoints to sort products by price and rating." \\
\textit{GEval = 70\%, $\Delta$ = –10\%} \\[4pt]
\textbf{What changed:} NC matches the GT’s client-side focus. With context, the model shifts to backend API work, plausible but more detailed than the intended abstraction.
\end{answerbox}

These examples show that architectural context can sharpen task generation but also shift the model focus away from simpler UI logic, depending on what the screen  conveys.

\subsection{Perception of Development Teams (RQ3)}
\label{subsec:teams-perc-results}
We summarise the teams' responses along the four questions introduced in Section \ref{subsec:teams-perc}.

\textbf{(1) Intended use of AI-generated backlogs.} 
All CapStone teams mentioned inspiration and ideation as primary reasons for using an AI backlog suggestion tool. 
Teams in Project 5 and Project 6 added that it could help when they ``have an idea but don’t know how to operationalise it.'' 
The team in Project 4 specifically mentioned using it for requirements review and validation, while the team in Project 5 noted it could provide a ``general overview'' but should not be ``trusted completely.'' 
AppComp teams were less sceptical and could see such a tool as part of their routine workflow. Teams in Project 1 and Project 3 indicated they would use it frequently, whereas the team in Project 2 suggested it would be used occasionally for major planning iterations or requirements changes.

\textbf{(2) Preferred timing in the project lifecycle.} Opinions varied on when to use GenAI for backlog creation. 
Most teams preferred using the tool during backlog refinement, noting that it helps identify missing items (Project 2) or refine tasks already in progress (Project 5, Project 1). 
Two teams also saw the project’s start as a good time for ideation (Project 6, Project 7). 
Two CapStone teams (Project 4, Project 5) expressed concern that introducing AI too early could bias the initial requirements and project direction.

\textbf{(3) Backlog item types most suited for AI support.} All CapStone teams agreed that scenarios/detailed user stories were the most challenging item to create and thus the most beneficial to automate. 
In contrast, AppComp teams found tasks (Projects 1 and 3) to be the most useful. 
A possible explanation is that AppComp teams rely more on tasks for planning and estimating work, and user stories often serve as a reference for project managers rather than developers. 
CapStone teams, on the other hand, may not have established processes for structuring backlog items, making automation of higher-level backlog items particularly valuable to them.

\textbf{(4) Perceived challenges and risks.} Commonly expressed concerns included model hallucinations (Project 7), irrelevant suggestions (Project 5), and unpredictability (Project 4). 
The team in Project 6 worried about over-reliance on AI, warning that it might discourage critical thinking \cite{rahe_how_2025}. 
Prompting difficulty and the need to ``take the AI’s answers with a grain of salt'' were also repeated themes. 
AppComp teams mentioned their main concern was getting redundant or unnecessary details in the generated items. 
Across all teams, developer oversight came up repeatedly. As one team noted, ``If you rely too much on it, you would not think for yourself. It should be used with caution, there must be a balance of human feedback and refinement."

\subsection{Ratio of Useful False Positives}
Table~\ref{tab:useful_fps} reports the proportion of false positives that teams chose to keep (as-is or with changes), averaged across AppComp and CapStone projects. Per-project breakdowns are in the replication package.

\begin{table}[htbp]
\centering
\caption{Average ratio of useful FPs by item  and prompt.}
\label{tab:useful_fps}
\small
\begin{tabular}{cl|l|l}
\hline
 & & \textbf{Avg. AppComp} & \textbf{Avg. CapStone} \\ \hline
\multirow{3}{*}{Epics} & Baseline & \mybar[bluebars]{3} & \mybar[pinkbars]{5} \\
 & CCoT    & \mybar[bluebars]{2} & \mybar[pinkbars]{4} \\
 & Persona & \mybar[bluebars]{1} & \mybar[pinkbars]{3} \\ \hline
\multirow{3}{*}{US/S} & Baseline & \mybar[bluebars]{4} & \mybar[pinkbars]{9} \\
 & CCoT    & \mybar[bluebars]{4} & \mybar[pinkbars]{5} \\
 & Persona & \mybar[bluebars]{4} & \mybar[pinkbars]{6} \\ \hline
\multirow{3}{*}{Tasks} & Baseline & \mybar[bluebars]{26} & \mybar[pinkbars]{9} \\
 & CCoT    & \mybar[bluebars]{20} & \mybar[pinkbars]{14} \\
 & Persona & \mybar[bluebars]{22} & \mybar[pinkbars]{7} \\ \hline
\end{tabular}
\end{table}

False positives were frequently considered useful, with tasks showing the highest rates (AppComp: 20--26\%; CapStone: 7--14\%). CapStone teams found more value in epic and scenario FPs, often because AI outputs split bundled human-written items into finer-grained entries. For example, one social-app scenario covering event joining, attendance confirmation, and photo upload was matched by three separate model items, each of which teams found useful. AppComp teams, who rely more on tasks for planning, showed a stronger preference for retaining task-level FPs.

\subsection{Revised Recall}
Standard recall penalises all false positives, assuming they provide no value. However, our analysis showed that many false positives still contributed by capturing missing interactions or structuring information differently from human-written items. This aligns with graded relevance approaches in information retrieval \cite{kanoulas} and with findings by Jiang et al.~\cite{jiang2024}, who report that LLM outputs diverging from a reference can still add value in open-ended tasks. In our context, such divergences correspond to “useful false positives”: items absent from the ground truth but retained or adapted by developers to address unrecorded needs or reframe existing requirements.

To reflect this, we suggest a Revised Recall (RR) measure as a complement to standard recall. RR incorporates the proportion of false positives judged useful by participants, allowing partially correct items to contribute to the score:
\begin{equation}
\text{RR} = \frac{\text{TP} + \alpha \cdot \text{UFP}}{\text{TP} + \text{FN} + \alpha \cdot \text{UFP}}
\end{equation}
UFP (Useful False Positives) represents false positives that were retained or modified, and \(\alpha\) is a weighting factor that controls how much they contribute to recall. 
We compute \(\alpha\) based on team's feedback, specifically:
\begin{equation}
\alpha = \frac{\text{Total ``Yes''} + 0.5 \cdot \text{Total ``Yes with Changes''}}{\text{UFP}}
\end{equation}
where Total ``Yes'' refers to false positives that were fully accepted, Total ``Yes with Changes'' refers to false positives that were retained but required modifications and Total UFP is the total set of useful false positives.
The \(0.5\) factor ensures that modified false positives are counted as partially correct, rather than fully correct. 
Higher values of \(\alpha\) indicate that most useful false positives were fully accepted, while lower values suggest that many required modifications, reducing their impact on recall. 
This follows prior discussions in graded relevance, where items may be partially correct or still beneficial despite not matching ground truth. Table~\ref{tab:revised_recall} presents the results across prompting strategies and backlog item types. Highlighted $\Delta$ values indicate cases where integrating useful false positives had the greatest impact on recall.

\begin{table*}[htbp]
\centering
\caption{Standard Recall (R) and Revised Recall (RR) across prompting strategies.}
\label{tab:revised_recall}
\resizebox{\textwidth}{!}{%
\begin{tabular}{@{}crrrrrrrrrrrrrrrrrrrrrrrrrrr@{}}
\cmidrule(l){2-28}
 & \multicolumn{9}{c}{\textbf{Baseline}} & \multicolumn{9}{c}{\textbf{CCoT}} & \multicolumn{9}{c}{\textbf{Persona}} \\ \cmidrule(l){2-28} 
 & \multicolumn{3}{c|}{Epics} & \multicolumn{3}{c|}{US/S} & \multicolumn{3}{c|}{Tasks} & \multicolumn{3}{c|}{Epics} & \multicolumn{3}{c|}{US/S} & \multicolumn{3}{c|}{Tasks} & \multicolumn{3}{c|}{Epics} & \multicolumn{3}{c|}{US/S} & \multicolumn{3}{c}{Tasks} \\ \cmidrule(l){2-28} 
 & \multicolumn{1}{c}{R} & \multicolumn{1}{c}{RR} & \multicolumn{1}{c|}{$\Delta$} & \multicolumn{1}{c}{R} & \multicolumn{1}{c}{RR} & \multicolumn{1}{c|}{$\Delta$} & \multicolumn{1}{c}{R} & \multicolumn{1}{c}{RR} & \multicolumn{1}{c|}{$\Delta$} & \multicolumn{1}{c}{R} & \multicolumn{1}{c}{RR} & \multicolumn{1}{c|}{$\Delta$} & \multicolumn{1}{c}{R} & \multicolumn{1}{c}{RR} & \multicolumn{1}{c|}{$\Delta$} & \multicolumn{1}{c}{R} & \multicolumn{1}{c}{RR} & \multicolumn{1}{c|}{$\Delta$} & \multicolumn{1}{c}{R} & \multicolumn{1}{c}{RR} & \multicolumn{1}{c|}{$\Delta$} & \multicolumn{1}{c}{R} & \multicolumn{1}{c}{RR} & \multicolumn{1}{c|}{$\Delta$} & \multicolumn{1}{c}{R} & \multicolumn{1}{c}{RR} & \multicolumn{1}{c}{$\Delta$} \\ \midrule
Project 1 
& 74\% & 86\% & \multicolumn{1}{r|}{{\color[HTML]{4EA72E} 12\%}} 
& 60\% & 79\% & \multicolumn{1}{r|}{{\color[HTML]{4EA72E} 19\%}} 
& 47\% & 85\% & \multicolumn{1}{r|}{{\color[HTML]{4EA72E} 37\%}} 
& 74\% & 84\% & \multicolumn{1}{r|}{{\color[HTML]{4EA72E} 10\%}} 
& 72\% & 83\% & \multicolumn{1}{r|}{{\color[HTML]{4EA72E} 11\%}} 
& 47\% & 76\% & \multicolumn{1}{r|}{{\color[HTML]{4EA72E} 28\%}} 
& 67\% & 79\% & \multicolumn{1}{r|}{{\color[HTML]{4EA72E} 13\%}} 
& 53\% & 68\% & \multicolumn{1}{r|}{{\color[HTML]{4EA72E} 14\%}} 
& 23\% & 70\% & \cellcolor[HTML]{DAF2D0} \textbf{{\color[HTML]{4EA72E} 47\%}} \\
Project 2 
& 78\% & 83\% & \multicolumn{1}{r|}{{\color[HTML]{4EA72E} 5\%}} 
& 67\% & 74\% & \multicolumn{1}{r|}{{\color[HTML]{4EA72E} 7\%}} 
& 61\% & 80\% & \multicolumn{1}{r|}{{\color[HTML]{4EA72E} 19\%}} 
& 68\% & 74\% & \multicolumn{1}{r|}{{\color[HTML]{4EA72E} 6\%}} 
& 67\% & 78\% & \multicolumn{1}{r|}{{\color[HTML]{4EA72E} 12\%}} 
& 53\% & 76\% & \multicolumn{1}{r|}{{\color[HTML]{4EA72E} 23\%}} 
& 72\% & 75\% & \multicolumn{1}{r|}{{\color[HTML]{4EA72E} 3\%}} 
& 53\% & 61\% & \multicolumn{1}{r|}{{\color[HTML]{4EA72E} 8\%}} 
& 46\% & 72\% & \cellcolor[HTML]{DAF2D0} \textbf{{\color[HTML]{4EA72E} 26\%}} \\
Project 3 
& 88\% & 92\% & \multicolumn{1}{r|}{{\color[HTML]{4EA72E} 4\%}} 
& 68\% & 79\% & \multicolumn{1}{r|}{{\color[HTML]{4EA72E} 11\%}} 
& 54\% & 84\% & \multicolumn{1}{r|}{{\color[HTML]{4EA72E} 30\%}} 
& 76\% & 82\% & \multicolumn{1}{r|}{{\color[HTML]{4EA72E} 6\%}} 
& 68\% & 79\% & \multicolumn{1}{r|}{{\color[HTML]{4EA72E} 12\%}} 
& 44\% & 79\% & \multicolumn{1}{r|}{{\color[HTML]{4EA72E} 35\%}} 
& 52\% & 63\% & \multicolumn{1}{r|}{{\color[HTML]{4EA72E} 11\%}} 
& 51\% & 70\% & \multicolumn{1}{r|}{{\color[HTML]{4EA72E} 19\%}} 
& 46\% & 81\% & \cellcolor[HTML]{DAF2D0} \textbf{{\color[HTML]{4EA72E} 35\%}} \\
\midrule
Project 4 
& 100\% & 100\% & \multicolumn{1}{r|}{0\%} 
& 80\% & 90\% & \multicolumn{1}{r|}{{\color[HTML]{4EA72E} 10\%}} 
& 50\% & 59\% & \multicolumn{1}{r|}{{\color[HTML]{4EA72E} 9\%}} 
& 100\% & 100\% & \multicolumn{1}{r|}{0\%} 
& 60\% & 75\% & \multicolumn{1}{r|}{\cellcolor[HTML]{DAF2D0} \textbf{{\color[HTML]{4EA72E} 15\%}}} 
& 36\% & 52\% & \multicolumn{1}{r|}{\cellcolor[HTML]{DAF2D0} \textbf{{\color[HTML]{4EA72E} 15\%}}} 
& 80\% & 86\% & \multicolumn{1}{r|}{{\color[HTML]{4EA72E} 6\%}} 
& 60\% & 75\% & \multicolumn{1}{r|}{\cellcolor[HTML]{DAF2D0} \textbf{{\color[HTML]{4EA72E} 15\%}}} 
& 41\% & 54\% & {\color[HTML]{4EA72E} 13\%} \\
Project 5 
& 50\% & 60\% & \multicolumn{1}{r|}{{\color[HTML]{4EA72E} 10\%}} 
& 100\% & 100\% & \multicolumn{1}{r|}{0\%} 
& 24\% & 37\% & \multicolumn{1}{r|}{{\color[HTML]{4EA72E} 12\%}} 
& 57\% & 63\% & \multicolumn{1}{r|}{{\color[HTML]{4EA72E} 5\%}} 
& 67\% & 75\% & \multicolumn{1}{r|}{{\color[HTML]{4EA72E} 8\%}} 
& 18\% & 38\% & \multicolumn{1}{r|}{\cellcolor[HTML]{DAF2D0} \textbf{{\color[HTML]{4EA72E} 21\%}}} 
& 38\% & 47\% & \multicolumn{1}{r|}{{\color[HTML]{4EA72E} 10\%}} 
& 67\% & 82\% & \multicolumn{1}{r|}{{\color[HTML]{4EA72E} 15\%}} 
& 20\% & 27\% & {\color[HTML]{4EA72E} 7\%} \\
Project 6 
& 20\% & 47\% & \multicolumn{1}{r|}{\cellcolor[HTML]{DAF2D0} \textbf{{\color[HTML]{4EA72E} 27\%}}} 
& 50\% & 75\% & \multicolumn{1}{r|}{{\color[HTML]{4EA72E} 25\%}} 
& 15\% & 19\% & \multicolumn{1}{r|}{{\color[HTML]{4EA72E} 4\%}} 
& 30\% & 50\% & \multicolumn{1}{r|}{{\color[HTML]{4EA72E} 20\%}} 
& 50\% & 70\% & \multicolumn{1}{r|}{{\color[HTML]{4EA72E} 20\%}} 
& 13\% & 20\% & \multicolumn{1}{r|}{{\color[HTML]{4EA72E} 8\%}} 
& 40\% & 43\% & \multicolumn{1}{r|}{{\color[HTML]{4EA72E} 3\%}} 
& 50\% & 70\% & \multicolumn{1}{r|}{{\color[HTML]{4EA72E} 20\%}} 
& 8\% & 15\% & {\color[HTML]{4EA72E} 7\%} \\
Project 7 
& 67\% & 80\% & \multicolumn{1}{r|}{{\color[HTML]{4EA72E} 13\%}} 
& 100\% & 100\% & \multicolumn{1}{r|}{0\%} 
& 18\% & 27\% & \multicolumn{1}{r|}{{\color[HTML]{4EA72E} 9\%}} 
& 100\% & 100\% & \multicolumn{1}{r|}{0\%} 
& 100\% & 100\% & \multicolumn{1}{r|}{0\%} 
& 13\% & 24\% & \multicolumn{1}{r|}{{\color[HTML]{4EA72E} 11\%}} 
& 67\% & 83\% & \multicolumn{1}{r|}{\cellcolor[HTML]{DAF2D0} \textbf{{\color[HTML]{4EA72E} 17\%}}} 
& 67\% & 83\% & \multicolumn{1}{r|}{\cellcolor[HTML]{DAF2D0} \textbf{{\color[HTML]{4EA72E} 17\%}}} 
& 23\% & 29\% & {\color[HTML]{4EA72E} 5\%} \\ \bottomrule
\end{tabular}%
}
\end{table*}

Across all cases, RR exceeds standard recall, with gains varying by item type. Tasks improve the most, suggesting that many task-level false positives still captured relevant work from the teams’ perspective. User stories and scenarios also benefit,
whereas epics tend to have rather small changes.

RR gains also differ by prompting strategy. Persona prompts yield the largest task increases (e.g., Project 1: \(\Delta = 47\%\), Project 3: \(\Delta = 35\%\)), while Baseline and CCoT vary more across projects. This suggests that prompt design affects how often false positives contain information developers consider relevant. Although Persona shows lower precision and F1 than Baseline and CCoT, its usefulness is clearer under RR: it produces larger gains, indicating that teams judged many of its false positives as useful. Rather than maximizing overlap with the ground truth, Persona favors breadth and user-centred phrasing, which can surface missing items and complement CCoT’s more implementation-aware outputs.

Overall, the consistent RR gains for tasks (\(\Delta\) often 19--47\%) indicate that developer feedback on false positives frequently revealed relevant content. Absolute RR for tasks often exceeded 80\% in AppComp projects, and ranged between 15\% and 59\% in CapStone projects. Incorporating this feedback provides a complementary view of AI-generated items, capturing contributions that ground-truth matching would discard.

\section{Discussion}
\label{sec:discussion}
\subsection{Towards Accurate Backlog Generation}
Because app mockups are easy to produce (e.g., Figma) and are often available early in requirements elicitation and negotiation, they offer a practical input for LLM-based backlog generation. Across projects, F1 for epics and (in most cases) user stories exceeded 50\% (e.g., 83\% in Project~2), while task-level performance was lower (15--45\%). 

Prompt design influences what types of items are generated and how accurate they are. Baseline favours recall over precision, surfacing many items from visible cues. CCoT adds intermediate reasoning via UI scene descriptions and can improve alignment for structured items such as epics and user stories, but for dense or domain-specific UIs the extra steps can divert attention. Persona, designed to emphasise user requirements, tends to underperform on technical, task-level items (e.g., Project 1 Tasks 16\% vs Baseline 31\% and CCoT 29\%) suggesting that a user-centric framing can reduce the technical specificity needed for detailed outputs. 

Architectural context has mixed effects. Brief stack descriptions help when tasks require cross-component coordination, but distract when the screen already encodes the relevant functionality. Context works best when it adds information not visible in mockups.

Our findings suggest that no single prompting approach works across all backlog levels; a hybrid, stepwise process is likely more effective. This study is a first step, given the heterogeneity of projects and backlogs. Future work should examine mockup characteristics (e.g., fidelity, style, UI density), project factors (e.g., backlog granularity and structure), and additional artefacts available at generation time (e.g., codebase, textual requirements, similar projects). Task types and granularity should be specified explicitly, but even minimal context must be handled with care as it can mislead the model.

\subsection{Model-Developers Interaction in Generative Requirements}
Our results have implications for how foundation models should support backlog creation. 
During early ideation, prompts with high recall (e.g., Baseline) may help uncover missing elements, whereas structured prompts (e.g., CCoT) yield more system-aligned drafts for epics and user stories that teams can refine.  
Architectural context should be introduced selectively: it benefits tasks involving backend logic, integration, or persistence, but may reduce clarity in UI-centred scenarios. Whether and when to introduce this context depends on the item type. It may be useful when behaviour is not visible in mockups, yet potentially overshadowing what screens already convey if overly detailed. Context granularity should match task scope~\cite{Ullrich2025ReqToCode}. Oversight is thus necessary to ``think in a different direction'' \cite{Wei:Software:2025}, filter hallucinations, and adapt AI outputs to team-specific conventions. 
Backlog creation is a creative and collaborative process \cite{Pham:ICSEW:2018, Wei:Software:2025} where developers, product owners, and potentially users discuss and iteratively refine items. 
AI-generated items then become starting points, material to discuss, edit, or discard~\cite{Halili2025Augmenting}. 
Differences between teams reflect this: CapStone valued ideation breadth, while AppComp prioritised consistency and workflow integration. Future work should study conversational, multi-turn generation that better mirrors how teams iteratively shape requirements.
   
\subsection{Incorporating Usefulness Judgements into Evaluation}
Our results show that exact-match metrics (such as Precision, Recall, and F1) did not always align with team judgments: a significant portion of “false positives” were retained because they were actionable, inspiring, or filled backlog gaps. 
This mirrors human--AI ideation work, where breadth and perceived usefulness drive value~\cite{shaer2024aiaugmentedbrainwriting}, and supports the view that backlog creation is inherently creative, with no single correct set of items~\cite{Wei:Software:2025, Pham:ICSEW:2018}.

We therefore introduce Revised Recall (RR) to complement exact-match metrics by counting items judged useful after review, even when they differ from the original ground truth. 
In our data, RR gains were most evident at the task level. 
This supports calls to move “beyond accuracy” in open-ended tasks and evaluate behaviour and utility, not only overlap \cite{ribeiro2020}. 
RR requires developer feedback collected after generation, which might limit its use as an evaluation metric at deployment time. 
The RQ3 procedure can nonetheless be reused for assessing AI outputs in other open-ended SE tasks, alongside standard exact-match measures.  Estimating RR without post-hoc interviews remains a direction for future work.

\subsection{Threats to Validity}
\label{subsec:tov}

\textbf{Internal and Conclusion Validity.}
Data leakage is a major threat when evaluating Foundation Models~\cite{kapoor_leakage_2023}; we mitigate it by studying seven new, non-public projects unseen during training.
Temperature was set to 0 for reproducibility; different settings or models could yield different results.
We iteratively refined the three prompts for clarity and release full details in the replication package.

\textbf{Construct Validity.}
Matching generated items to ground truth is partly subjective.
We used predefined criteria and peer-labelling by two authors; agreement with an independent annotator on a stratified subsample was high ($\kappa = 0.87$), and an LLM judge showed moderate alignment ($\kappa \approx 0.60$). Regarding usefulness judgements, ratings were collected in a single interview session and may reflect novelty effects rather than sustained practical value; moreover, the $\alpha$ weighting in RR is derived from the same participants, so RR values should be interpreted as team-specific rather than externally validated.

\textbf{External Validity.}
We studied seven real projects across different domains (e-commerce, event planning, recipe assistance, meal management) and two organisational settings (AppComp and CapStone), covering iOS, Flutter, Swift, and web stacks. Replication on larger systems, additional domains, and other environments is needed. Architectural context descriptions in RQ2 were also not standardized across projects, which limits direct comparison of context effects. We used a single model (GPT-4o); although performance varies across model families, surveys report narrowing gaps among frontier models~\cite{aiindex2025}, so we expect similar relative trends, even if absolute scores differ.

\section{Related Work}
\label{sec:relwork}
Generative AI has been applied across SE tasks~\cite{Fan2023, Hou2024}.
In requirements engineering, LLMs support elicitation~\cite{Wei2024}, specification~\cite{ma2025spec, xie2025}, and user story generation~\cite{Oswal2024, Rahman2024}.
White et al.~\cite{white2023c} and Vogelsang and Fischbach~\cite{Vogelsang2024a} provide prompt engineering guidelines and systematic guidance for applying LLMs across RE tasks more broadly.
Recent work has also examined structured elicitation interviews with LLM-based agents~\cite{Korn2025LLMREI} and iterative chat-based refinement~\cite{wang2023}.
Zhang et al.~\cite{Zhang2023} applied LLMs to zero-shot requirements retrieval, achieving high recall but low precision, a trade-off we also observe.
LLMs have also been explored for feature ideation: Wei et al.~\cite{Wei2024} found that LLMs can suggest novel sub-features beyond existing market offerings, though feasibility checks are needed.
El-Hajjami et al.~\cite{El-Hajjami2023} and Kutzner et al.~\cite{Kutzner2023} both highlight the importance of human oversight to mitigate hallucinations; a concern our participants also raised.

Backlog generation has received less attention. 
Oswal et al.~\cite{Oswal2024} transformed textual requirements into user stories using few-shot prompting, and Rahman and Zhu~\cite{Rahman2024} introduced GeneUS with a ``Refine and Thought'' strategy. Both rely on textual inputs; our work uses multimodal models to translate visual mockups into full backlog hierarchies (epics, user stories, and tasks).

Vision--language models (VLMs) combine textual and visual understanding, making them relevant for SE tasks involving GUIs. Applications include UI search~\cite{Wei2024b}, usability evaluation~\cite{pourasad2024}, test generation~\cite{Liu:MakeLLMTesting:2024, Wang:ICSME:25}, and UI code generation from screenshots~\cite{Wan2024}. None of these studies examined backlog generation from visual inputs; our work extends VLM applications to early-stage mockup-to-backlog translation.

\section{Conclusion}
\label{sec:conclusion}
In this paper, we studied the use of GenAI for generating sprint backlogs from visual mockups. 
Across seven app projects (110 mockups, 595 backlog items), we compared Baseline, CCoT, and Persona prompts using quantitative metrics and team feedback. 
Baseline achieved high recall for epics and user stories in structured backlogs (73--85\%) but lower precision, particularly for user stories ($\approx$49--61\%). CCoT often improved epics but was mixed for user stories and tasks, and Persona generally underperformed on tasks. 
Adding brief architectural context improved task-level performance, particularly for backend work, though effects varied by prompt and project. 
To capture practical value, we introduced Revised Recall (RR), which counts team-judged useful outputs beyond exact matches and proved most impactful at task level. 
Our findings support the role of LLMs as collaborative assistants in early requirements and planning work, provided their outputs are reviewed by developers. 
Future work should test RR in other open-ended SE tasks and examine how context should be calibrated to project needs.

\section*{Data Availability}
We provide the templates and analysis scripts in the replication package.
Due to business and privacy agreements, the actual backlogs and detailed app mockups cannot be publicly released. 

\balance
\bibliographystyle{ieeetr}
\bibliography{bibliography}

\end{document}